%% file: 0-main.tex
\documentclass[sigconf,nonacm]{acmart}

\usepackage{times}
\usepackage{soul}
\usepackage{url}
\usepackage{subcaption}
\usepackage{xcolor}

\usepackage{amsmath,bm}
\usepackage{multirow}

\newcommand{\commentout}[1]{%
}

\newcommand{\fixme}[1]{{\textcolor{red}{\textit{#1}}}}
\newtheorem{problem}{Problem}

\title[Cold Causal Demand Forecasting Model]{Mitigating Cold-start Forecasting using \\ Cold Causal Demand Forecasting Model}

\author{Zahra Fatemi}
\authornote{Work done during internship with Google Cloud}
\email{zfatem2@uic.edu}
\affiliation{%
  \institution{University of Illinois Chicago}
  \country{}
}
\author{Minh Huynh}
\email{mahuynh@google.com}
\affiliation{%
  \institution{Google Inc.}
  \country{}
}
\author{Elena Zheleva}
\email{ezheleva@uic.edu}
\affiliation{%
  \institution{University of Illinois Chicago}
  \country{}
}
\author{Zamir Syed}
\email{zamirs@google.com}
\affiliation{%
  \institution{Google Inc.}
  \country{}
}
\author{Xiaojun Di}
\email{xdi@google.com}
\affiliation{%
  \institution{Google Inc.}
  \country{}
}
\setcopyright{none}
\renewcommand\footnotetextcopyrightpermission[1]{}
\begin{document}

\begin{abstract}

Forecasting multivariate time series data, which involves predicting future values of variables over time using historical data, has significant practical applications. Although deep learning-based models have shown promise in this field, they often fail to capture the causal relationship between dependent variables, leading to less accurate forecasts. Additionally, these models cannot handle the cold-start problem in time series data, where certain variables lack historical data, posing challenges in identifying dependencies among variables. To address these limitations, we introduce the \textit{Cold Causal Demand Forecasting (CDF-cold)} framework that integrates causal inference with deep learning-based models to enhance the forecasting accuracy of multivariate time series data affected by the cold-start problem.
To validate the effectiveness of the proposed approach, we collect 15 multivariate time-series datasets containing the network traffic of different Google data centers. Our experiments demonstrate that the CDF-cold framework outperforms state-of-the-art forecasting models in predicting future values of multivariate time series data.

\end{abstract}
\maketitle
\input{1-introduction}
\input{2-related}
\input{3-problem}
\input{4-solution}

\input{5-experiments}

\input{6-conclusion}

\bibliographystyle{ACM-Reference-Format}
\bibliography{references}
\end{document}

%% file: 1-introduction.tex
\section{Introduction}
Time series forecasting plays a central role in a vast number of studies \cite{wu-icml13, martinez-eng15, moorthy-tpt88, kaushik-fro20}. The goal of time series forecasting is to use historical and current data to predict future values over a period of time. The applications of time series prediction are diverse and include climate and weather forecasting in geography \cite{dastorani-naha16,ghil-rog02}, traffic flow prediction in transportation \cite{xu-arxiv20,cirstea-icde22,yu-jcai18}, healthcare diagnosis in medical science \cite{soyiri-ehp13,bui-bme18}, and sales and stock prices prediction in economics \cite{kim-neu03,mondal-jcse14,krollner-esa10,cao-jcs19}. We specifically focus on forecasting the network traffic of different Google data centers as a motivating example. Each data center hosts multiple Google services with different network traffic and machine usage. Accurately forecasting the network traffic in these data centers is critical for efficient resource allocation and capacity planning, as well as ensuring a high-quality user experience.

In recent years, there has been extensive research and application of various time series forecasting methods. While classic forecasting methods like Univariate Autoregressive (AR), Univariate Moving Average (MA), Simple Exponential Smoothing (SES), and Autoregressive Integrated Moving Average (ARIMA) \cite{box-jws15} have been widely studied, they are limited by their assumptions of linearity and aperiodicity of data. Moreover, these models fail to effectively forecast multivariate time series datasets, where each variable's behavior depends not only on its past values but also on the interactions with other variables. Recently, deep learning models have been developed to capture the complexity and nonlinearity in time series forecasting. Long Short-Term Memory (LSTM) is one of the prominent deep learning models used to extract dynamic information from time series data through the memory mechanism \cite{sak-int14,siami-icmla18}.

While these approaches often perform well at capturing temporal patterns, they often overlook the interdependencies between different time series variables. The better the interdependencies among different time series are modeled, the more accurate the forecasting can be \cite{xu-arxiv20}. 
In the running example, the impact of different Google services on the network traffic can vary depending on the machine usage of the service.
Recently, Graph Neural Networks (GNN) have been utilized to incorporate the topology structure and interdependencies among variables in forecasting tasks \cite{wang-arxiv22,yu-jcai18}.  In GNN models, each variable from a multivariate time series is represented as a node in a graph, and the edges between the nodes capture the interdependencies or relationships between the variables. 
By propagating information between neighboring nodes, Graph Neural Networks (GNNs) empower each variable in a multivariate time series to be aware of the influence of correlated variables.  However, these models often lack the ability to effectively capture and understand causal relationships between variables.
Incorporating causal knowledge into forecasting models enhances interpretability and provides insight into the factors that affect the target variable. 
Previous studies have attempted to quantitatively characterize the interdependencies between time series variables through causality \cite{kirchgassner-spr13}. Granger causality is one commonly used approach in time series analysis, particularly in economics \cite{granger-granger69}. Nevertheless, research has demonstrated that Granger causality cannot handle nonlinear relationships well, often leading to spurious causality or the identification of false causal relationships.
\cite{cartwright_cam2007}.

While it is straightforward to train a forecasting model with past values of the time series, forecasting time series data with no historical data, known as cold-start forecasting, is challenging.  In the absence of historical data, the forecasting model fails to capture and learn the inter-dependencies between new and existing variables, leading to inaccurate predictions of the target variables. 
For instance, when new Google services are introduced to a data center in the future, they may impact the total network traffic of the data center.

Motivated by applications in Google data centers, we propose the Cold Causal Demand Forecasting (CDF-cold) framework that brings together causal inference and deep learning-based models to increase the accuracy of multivariate time series forecasting suffering from the cold-start problem. CDF-cold consists of two main components: 1) The Causal Demand Forecast component that exploits the causal relationship between different variables of a multivariate time series to learn a new representation for each variable based on the representation of other variables that causally impact it, and 2) The Cold-start forecasting component that leverages similarity-based approaches to alleviate the cold-start forecasting in time series data. While the idea of using deep learning models in time series forecasting is not novel, combining these models with causal inference to address the cold-start problem in multivariate time series forecasting is novel.
To summarize, this paper makes the following contributions:
\begin{itemize}
\item We formulate the cold-start forecasting problem in multivariate time series datasets. In particular, we focus on datasets where future values for some variables correlated with the variables with no historical data are available in advance.
\item We develop a similarity-based framework that incorporates causal relationships between variables in a deep learning model and addresses cold-start forecasting in time series data with no historical data.
\item  We evaluate the performance of our framework on 15 multivariate time series datasets from various Google data centers, containing network traffic, and machine usage of different Google services. Our results demonstrate that our proposed framework outperforms existing baselines in forecasting accuracy.
\end{itemize}

The rest of the paper is structured as follows. In Section 2, we provide a brief overview of related work in time series forecasting.
In Section 3, we introduce the preliminaries and the problem that we address in the paper.
In Section 4, we present our Cold Causal Demand Forecasting framework.
In Section 5, we present our experimental setup and results.
Finally, in Section 6, we conclude and discuss directions for future work.


%% file: 2-related.tex
\section{Related Work}
In this section, we provide an overview of the classical and deep learning-based forecasting models, as well as recent advancements in causal-based forecasting.

\textbf{Classical models for forecasting}:
Classical forecasting methods rely on statistical regression techniques to predict future values of variables based on historical information. The Autoregression (AR) method models the next time points in a time series as a linear function of the observations at prior time steps \cite{sullivan-jss94,kashyap-ieee82}. The Moving Average (MA) method models the next time points in a time series as a linear function of the residual errors from a mean process at prior time steps \cite{kashyap-ieee82}. Auto Regressive Moving Average (ARMA), which combines both AR and MA models, is a class of models that forecasts future values of a given time series as a linear function of the observations and residual errors at prior time points \cite{chen-eps95}. The Autoregressive Integrated Moving Average (ARIMA) is one of the most widely used statistical methods for time series forecasting in non-seasonal time series that exhibit non-random patterns. It combines AR, MA, and a differencing pre-processing step called integration (I) to make the time series stationary \cite{ho-els98,box-jws15}. In Vector Auto Regression (VAR), each variable is a linear function of its past values and the past values of all the other variables. Despite their popularity, these models cannot capture nonlinear and complex temporal patterns among different time series, resulting in sub-optimal forecasts.

\textbf{Deep Learning Based Forecasting}:
Recent studies have shown that deep learning methods can consistently outperform classical methods in time series forecasting \cite{li-iclear17,wang-icml19,hewamalage-ijf21}. A line of research focuses on Recurrent Neural Network (RNN)-based architectures for temporal forecasting applications. \citet{salinas-ijf20} propose DeepAR, a method for obtaining accurate probabilistic forecasts using an autoregressive RNN model on a large number of time series datasets. \cite{rangapuram-neurips18} develop a probabilistic time series forecasting that combines state space models with deep (recurrent) neural networks. \citet{madan-iccc18} propose an RNN-based technique for forecasting computer network traffic. Long Short-Term Memory (LSTM), which is a special type of RNN with additional features to memorize sequences of data, has gained lots of attention in recent years \cite{hochreiter-97nc}. \citet{zhao-iet17} develop an LSTM-based model to improve short-term traffic forecasting. To increase the speed of forecasting models and reduce the number of model parameters, \citet{yu-ijc18} propose a spatio-temporal graph convolutional network model for traffic forecasting tasks. Recent studies show that existing deep learning methods fail to fully exploit latent spatial dependencies between pairs of variables in multivariate time series forecasting. A new line of research focuses on deploying Graph Neural Networks (GNNs) to integrate the dependency between variables in the forecasting model and improve the prediction accuracy \cite{wu-kdd20,shang-iclr21,cheng-pr22,li-kbs19}. Recently, attention mechanisms (e.g., transformers) have shown superior performance in time series forecasting due to their ability in handling long-term dependencies \cite{de2-asc20,du-neu20}. \citet{cirstea-icde22} develop an attention-based method to enable spatio-temporal aware attention that is capable of modeling complex spatio-temporal dynamics in traffic time series datasets.

\textbf{Causal-based forecasting}: While there has been extensive research on developing deep learning methods for time series forecasting, less attention has been paid to the impact of causal relationships between variables in a multivariate time series on forecasting accuracy. However, identifying causal relationships in time series data remains a challenge. One prominent approach to identifying causal relationships is Granger causality \cite{granger-eco96,freeman-ajp83,chen-pl04,eichler-je07}, which examines whether the prediction of one time series can be improved by incorporating information from another time series. However, prior studies have shown that Granger causality may lead to spurious or falsely detected causal relationships due to its inability to handle nonlinear relationships \cite{cartwright_cam2007}. \citet{xu-arxiv20} develop an approach to identify causal relationships among variables and use this information in multivariate time series forecasting. However, this method does not address the cold-start forecasting problem.
Another recent study has proposed a method for causal discovery and forecasting in nonstationary time series data \cite{huang-icml19}. This work focuses on learning causal graphs from nonstationary time series data, which is an important step toward accurate forecasting of real-world scenarios.
Overall, these prior works have made significant contributions to the field of causal-based time series forecasting, but there is still room for improvement in identifying causal relationships and addressing the cold-start forecasting problem. Our proposed method aims to fill this gap by utilizing a novel causal-based approach that is capable of handling nonlinear relationships and improving the accuracy of forecasting for multivariate time series data suffering from cold-start problem. 

%% file: 3-problem.tex
\section{Preliminaries}
\label{preliminaries}
Suppose that we have $N$ data centers, and each data center $L_i \in \mathbf{L}$ generates a multivariate time series recording $A$ attributes (such as network traffic and machine usage of each Google service) over time. 
Let $\mathbf{X} \in \mathbb{R}^{N \times T \times A}$ denote the multivariate time series from all $N$ data centers for a total of $T$ time steps. We use $x_i \in \mathbb{R}^{T \times A}$ to represent the multivariate time series from data center $L_i$, $x_{i,T}$ to show the attributes from data center $L_i$ at time $T$, and $x^{j}_{i,t}$ to denote the value of attribute $a_j$ in data center $L_i$ at time $t$.

The goal of time series forecasting is to learn a function $F_{\theta}$ that, at timestamp t, given the attributes of the past $U$ timestamps from each data center $L_i$, predicts the values of the attributes in the future $H$ timestamps. Formally, at timestamp $t$, the forecasting function $F_{\theta}$ predicts the values of all variables in data center $L_i$ over the next $H$ timestamps as:
\begin{align}
(\hat{x}_{i,t+1},\hat{x}_{i,t+2},...,\hat{x}_{i,t+H})= 
    F_{\theta}(x_{i,t-U+1},x_{i,t-U+2},...,x_{i,t})
\end{align}
where $\theta$ is a set of learnable model parameters of the forecasting model, $H$ is the horizon ahead of the current timestamp, and $\hat{x}_{i,t-H+1}$ is the prediction for all variables in data center $i$ at time $t-H+1$.

\subsection{Cold-start forecasting problem}
Cold-start forecasting problem has posed a significant challenge in the field of forecasting. This issue arises when there is no historical data available for time series data or when the available data is insufficient to make reliable predictions.
The problem becomes even more complex when there are interdependencies between the time series impacted by the cold-start problem and other variables.
One example of such interdependencies is when introducing a new Google service to a data center, which can have a substantial impact on the overall network traffic within the data center. This sudden change in the network traffic pattern can disrupt the existing relationships and correlations that machine learning models rely on for forecasting. As a result, accurately predicting future patterns becomes more challenging.

Conventional machine learning and neural network forecasting models face challenges when attempting to derive precise inferences from inadequate information. These models often assume that all variables have an equal impact on the target variable, which does not always hold true in real-world domains. For example, highly used services can exert a more substantial influence on the total network traffic of a data center compared to less utilized services. This discrepancy in impact can lead to inaccuracies in forecasting when conventional models are employed.

In some cases, future information for certain attributes that are correlated with time series impacted by the cold-start problem may be available in advance (e.g., machine usage data for a new Google service). By leveraging this information in the forecasting model, it is possible to improve the accuracy of predictions and mitigate the cold-start problem to some extent.

\commentout{
\begin{align}
x_i=
\begin{bmatrix}
x^{(1)}_{i,1}&\emptyset&...&x^{(|A|-1)}_{i,1}&x^{(|A|)}_{i,1}\\
x^{(1)}_{i,2}&\emptyset&...&x^{(|A|-1)}_{i,2}&x^{(|A|)}_{i,2}\\
...&...&...&...\\
x^{(1)}_{i,T}&\emptyset&...&x^{(|A|-1)}_{i,T}&x^{(|A|)}_{i,T}
\end{bmatrix}
\end{align}}
For example, assume that in data center $L_i$, the historical data for attributes $a_2$ and $a_{A-1}$ is not available but the future values of attribute $a_{A-1}$ which is correlated with $a_2$ are available. Then, the forecasting task would be:
\begin{flalign}
\label{eq:example}
\begin{bmatrix}
\hat{x}^{(1)}_{i,t+1}&\hat{x}^{(2)}_{i,t+1}&...&x^{(A-1)}_{i,t+1}&\hat{x}^{(A)}_{i,t+1}\\
\hat{x}^{(1)}_{i,t+2}&\hat{x}^{(2)}_{i,t+2}&...&x^{(A-1)}_{i,t+2}&\hat{x}^{(A)}_{i,t+2}\\
...&...&...&...\\
\hat{x}^{(1)}_{i,t+H}&\hat{x}^{(2)}_{i,t+H}&...&x^{(A-1)}_{i,t+H}&\hat{x}^{(A)}_{i,t+H}
\end{bmatrix} 
=&
\\ F_\theta(\begin{bmatrix}
x^{(1)}_{i,t-U+1}&\emptyset&...&\emptyset&x^{(A)}_{i,t-U+1}\\
x^{(1)}_{i,t-U+2}&\emptyset&...&\emptyset&x^{(A)}_{i,t-U+2}\\
...&...&...&...\\
x^{(1)}_{i,t}&\emptyset&...&\emptyset&x^{(A)}_{i,t}\notag\\
-&-&...&x^{(A-1)}_{i,t+1}&-\\
...&...&...&...&...\\
-&-&-&x^{(A-1)}_{i,t+H}&-
\end{bmatrix})
\end{flalign}

In this paper, our goal is to predict ($\hat{x}_{i,t}, \hat{x}_{i,t+1}, ..., \hat{x}_{i,t+H}$) as accurate as possible. More formally:

\begin{problem}[cold-start forecasting]
\label{problem-def} 
 Given N multivariate time series corresponds to N data centers with A attributes in T timestamps represented by $\mathbf{X} \in R^{L \times T \times A}$, along with a forecasting horizon of H, we aim to learn a function $F_\theta$ that predicts the future values of attributes at each data center such that: 
\begin{equation}
\label{objective}
\begin{aligned}
& \underset{}{\mathrm{argmin}}
& \sum_{h=1}^H{\sum^{A}_{j=1}{(\hat{x}^j_{i,t+h}-x^j_{i,t+h})^2}}.
\end{aligned}
\end{equation}
\end{problem}

\commentout{
\begin{align}
x_i=
\begin{bmatrix}
x^{(1)}_{i,1}&\emptyset&...&x^{(|A|-1)}_{i,1}&x^{(|A|)}_{i,1}\\
x^{(1)}_{i,2}&\emptyset&...&x^{(|A|-1)}_{i,2}&x^{(|A|)}_{i,2}\\
...&...&...&...\\
x^{(1)}_{i,T}&\emptyset&...&x^{(|A|-1)}_{i,T}&x^{(|A|)}_{i,T}
\end{bmatrix}
\end{align}

\begin{align}
x_i=
\begin{bmatrix}
x^{(1)}_{i,1}&x^{(2)}_{i,1}&...&x^{(|A|-1)}_{i,1}&x^{(|A|)}_{i,1}\\
x^{(1)}_{i,2}&x^{(2)}_{i,2}&...&x^{(|A|-1)}_{i,2}&x^{(|A|)}_{i,2}\\
...&...&...&...\\
x^{(1)}_{i,T}&x^{(2)}_{i,T}&...&x^{(|A|-1)}_{i,T}&x^{(|A|)}_{i,T}
\end{bmatrix}
\end{align}
}

%% file: 4-solution.tex
\section{Cold Causal Demand Forecasting Model}
\label{solution}

With the aim of addressing the cold-start problem in multivariate time series forecasting, we propose \textit{Cold Causal Demand Forecasting (CDF-cold)} framework, which brings together causal inference and neural networks to improve forecasting accuracy for multivariate time series datasets suffering from cold-start problem. In this section, we provide a detailed description of the architecture of the CDF-cold framework.
\begin{figure*}[ht]
  \centering
  \includegraphics[width=\textwidth]{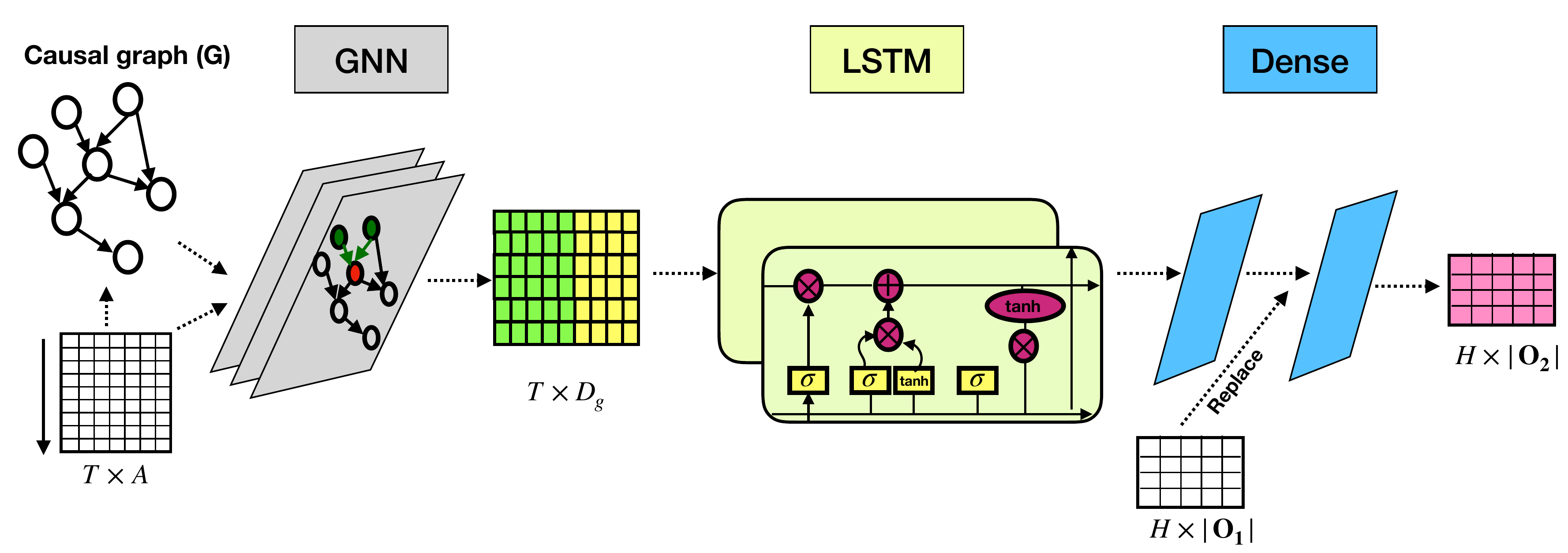}  
  \caption{Illustration of Causal Demand Forecast (CDF) component. \textbf{Input}: a multivariate time series dataset and the causal graph representing the cause and effect relationship between variables. \textbf{GNN}: generates a representation for each variable in each time point base on the variables causally impacting it.  
  \textbf{LSTM}: generates a representation for each variable based on the historical data and the representation generated by the GNN layer. 
  \textbf{Dense}: generate the forecasting for H horizons.
  }
  \label{fig:diag}
\end{figure*}
CDF-cold comprises two main components: 1) Causal Demand Forecasting, and 2) Cold-start Forecasting.

\subsection{Causal Demand Forecasting (CDF)}
The CDF component is designed to train a forecasting model for datasets with historical data for all variables. CDF consists of two sub-components:

\begin{itemize}
\item \textbf{Causal Component}. Existing time series forecasting models often assume correlations between a time series and its lags \cite{arya-ser15}. However, the future values of a time series can be influenced not only by its historical data but also by other variables in the dataset. Causal inference can play a crucial role in time series forecasting by identifying the causal relationships between a time series and other variables in the dataset, improving our understanding of how different factors affect the time series over time, and helping us make more accurate predictions about future values.

In our framework, we leverage causal inference to capture interdependencies between different variables.
To identify the causal relationships between variables in the multivariate time series dataset, we utilize existing causal discovery algorithms and integrate the resulting causal graph into the forecasting model. Structural Causal Models (SCMs) provide graphical representations of cause-effect relationships between variables that allow for the identification of the effect of interest \cite{pearl-book09}. Causal Graphs are Directed Acyclic Graph (DAG) representations of SCMs, where the direction of an edge determines the relationship between the two variables. If the variable Y is the child of a variable X, then we say that Y is caused by X, or that X is the direct cause of Y.

\item \textbf{Representation Learning Component}.
The goal of this component is to learn a representation for each time series based on lags of the time series and variables causally impacting it. This component is a multilayer neural network model consisting of:
\begin{itemize}
    \item Graph Neural Network layer: For representation learning based on other attributes, we leverage \textit{Graph Neural Networks (GNN)} whose effectiveness has been demonstrated in various machine learning tasks \cite{zhou-ai20,wu-kdd20,shang-iclr21}.
    GNN is a deep learning approach for semi-supervised learning on graph-structured data which gets the matrix representation of the graph structure and the attribute matrix as the input and generates a new representation for each node based on the attributes of its neighbors. 
    In our setting, for each data center $L_i$, We feed the causal graph G, extracted by the Causal Component, and the multivariate time series $x_i$ into the GNN layer to obtain a new representation for each target variable based on the variables causally impacting it within the causal graph as:
    \begin{align}
        h_{i}=\sigma((x_iM)\mathbf{W^1_1})
    \end{align}

\commentout{\begin{align}
        h_{i}=\sigma(\sigma(
\sigma((x_iM)\mathbf{W_1})
\mathbf{W^2_1})...W^s_1)
    \end{align}
    }
    where $M \in R^{A \times A}$ is the adjacency matrix corresponding to the causal graph G, $\mathbf{W_1}\in R^{T \times A \times D_g}$ is the
weight matrix to be learned, $D_g$ denotes the dimension of the new representation generated by GNN and $\sigma$ stands for the ReLU activation function. By applying the GNN layer to the input multivariate time series and the causal model, an attribute's representation is informed by the information from its neighbors. However, we need not only information from the causally impacting attributes but also we need to process the information of each attribute over time. With this goal, we pass each attribute's new representation through a recurrent layer. 
    \item Long Short-Term Memory Networks (LSTM): LSTM is a Recurrent Neural Network (RNN) model with the capability of memorizing the important parts of the input sequence seen so far for the purpose of future use \cite{hochreiter-lstm97}. The LSTM layer enables the model to memorize historical information for each time series and generate a new representation based on the representation generated by the GNN layer. For a model with s LSTM layers, we have:
    \begin{align}
        \hat{h}_i=\sigma(...\sigma(
        \sigma(h_{i} \mathbf{W^1_2})\mathbf{W^2_2})...W^s_2)
    \end{align}
    where $\mathbf{W^s_2}\in R^{ A \times D_l}$ and $D_l$ denotes the dimension of the new representation generated by the LSTM layer s.
    \item{Dense layer}: The output of LSTM is passed to a densely connected neural network layer to generate the forecasting for H horizons as:
    \begin{align}
        h^\prime_i=\sigma(\hat{h_i} \mathbf{W_3})
    \end{align}
    where $\mathbf{W_3}\in R^{ A \times H}$ represents the learnable weights for the dense layer. The output of the dense layer shows the forecasting for A attributes in time steps $t+1, t+2,...,t+H$.
\end{itemize}
Since we assume that the future values for some attributes are available in advance, we take advantage of this information to improve the forecasting of other time series. Let $\mathbf{O_1}$ denote the set of attributes with known future values $x^{|\mathbf{O_1}|}_{i,H} \in R^{H \times |\mathbf{O_1}|}$  in data center $L_i$ and let $\mathbf{O_2} $ represents all the attributes in A which are not in $\mathbf{O_1}$.
We concatenate the forecasting of set $\mathbf{O_2}$ in $h^\prime_i$ with $x^{\mathbf{O_1}}_{i,H}$ and obtain a new vector $\tilde{h_i}$ as:
\begin{align}
\tilde{h_i}=concat({h^\prime}_{i}^{|\mathbf{O_2}|},x^{|\mathbf{O_1}|}_{i,H}).
\end{align}
Then, we feed $\tilde{h_i}$ to a densely connected neural network layer to obtain the final forecasting:
\begin{align}
        \ddot{h}_i=\sigma(\tilde{h_i} \mathbf{W_4}),
    \end{align}
where $\mathbf{W_4}\in R^{ |\mathbf{O_2}| \times H}$ is the wight matrix to be learned. Fig. \ref{fig:diag} demonstrates the network architecture of the CDF component. \\
\end{itemize}

\subsection{Cold-start forecasting component} The utilization of the Representation Learning component enables the prediction of future values in time series datasets that have access to historical data. However, when faced with the cold-start problem, the model becomes insufficient in capturing the interdependencies between the target variable and variables without historical data. To address this issue, we propose a similarity-based approach that 
leverages forecasting models trained on the k most similar data centers with historical data. This component is comprised of three main steps as described below:
\begin{enumerate}
    \item We use a similarity-based approach to find k most similar data centers to the target data center with the cold-start problem. A data center can be considered  as a potentially similar data center if it has enough historical data for the variables with the cold-start problem in the target data center.
    \item We use the forecasting models trained on the k similar data centers to predict the future values of attributes in the target data center with no historical data. The input to each trained model would be the available future data for set $\mathbf{O_1}$ of the target data center $x_{i,H}^{|\mathbf{O_1}|}$.
    \item We take the average of the predictions for target variables of the target data center predicted by the models trained on the k similar data centers.
\end{enumerate}

We consider two different similarity-based approaches in our framework:
\begin{itemize}
    \item Gaussian Mixture Model (GMM): 
    GMM is a clustering technique that assumes a specific number of Gaussian distributions in the data, where each distribution represents a cluster \cite{reynolds-enc09}. By applying GMM to multiple multivariate time series datasets, we can group time series belonging to a single distribution together. The parameters of the GMM model are estimated using the Expectation-Maximization algorithm based on maximum likelihood.
\item Extended Frobenius norm (Eros): Eros is a Principal Component Analysis (PCA) based approach that measures the pairwise similarity between multiple multivariate time series datasets \cite{yang-md04}. In this approach, the covariance matrices of different datasets are measured. Then, the similarity between eigenvectors weighted by eigenvalues of the covariance matrices quantifies the similarity between different multivariate time series datasets.
\end{itemize}

%% file: 5-experiments.tex
\section{Experiments}
\label{experiments}
In this section, we evaluate the performance of different methods in multivariate time series forecasting. We first describe the dataset used in our experiments and then discuss our baselines and results. 
\subsection{Dataset}
We collect the multivariate time series of 15 Google data centers. Each dataset comprises network traffic and machine usage information for the top 200 Google services in terms of network traffic, as well as the overall network traffic for each data center, over a period of 533 days.
\begin{figure*}[ht]
\begin{subfigure}{0.49\textwidth}
  \centering
  \includegraphics[width=\textwidth]{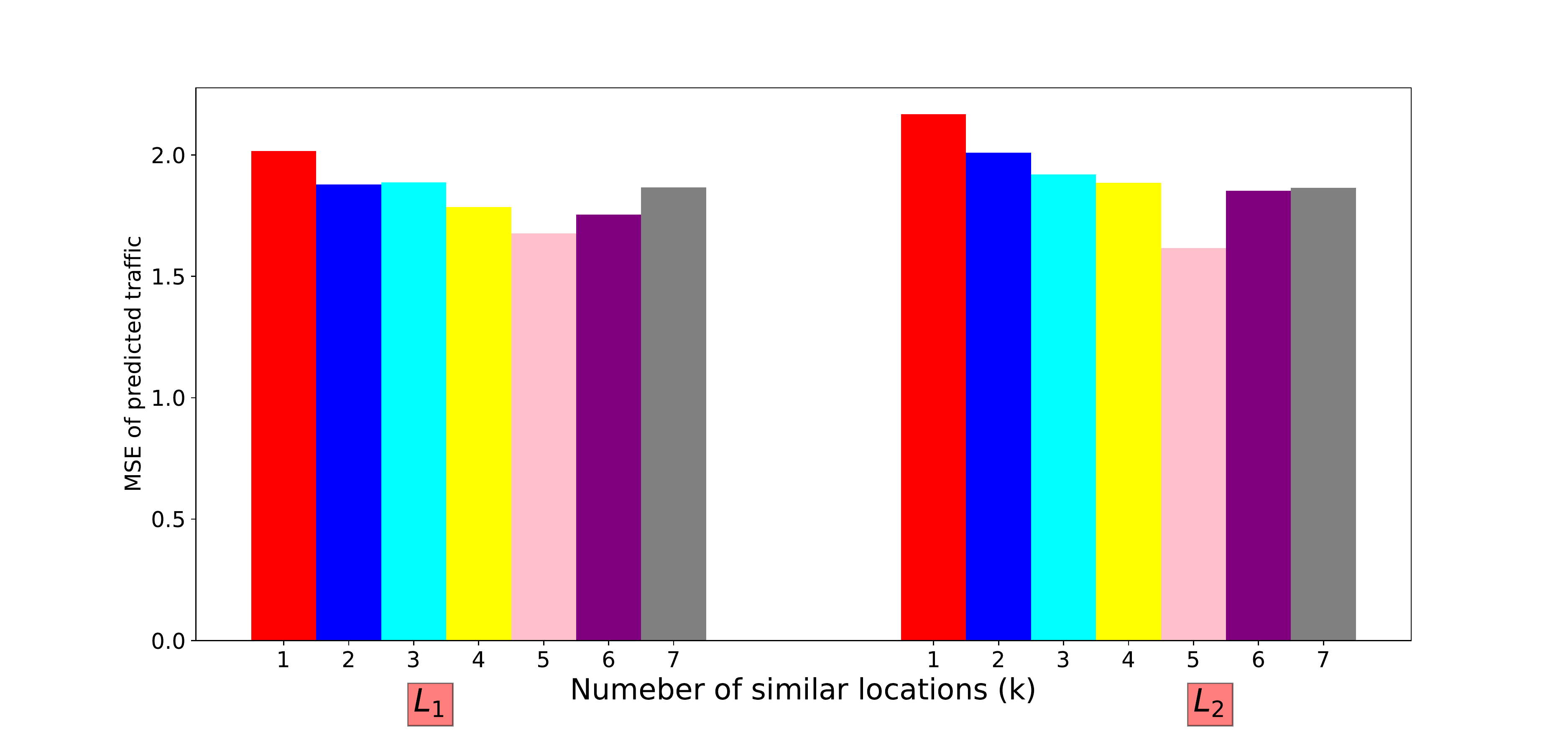}  
  \caption{CDF\_eros}
  \label{fig:h1}
\end{subfigure}
\begin{subfigure}{0.49\textwidth}
  \centering
  \includegraphics[width=\textwidth]{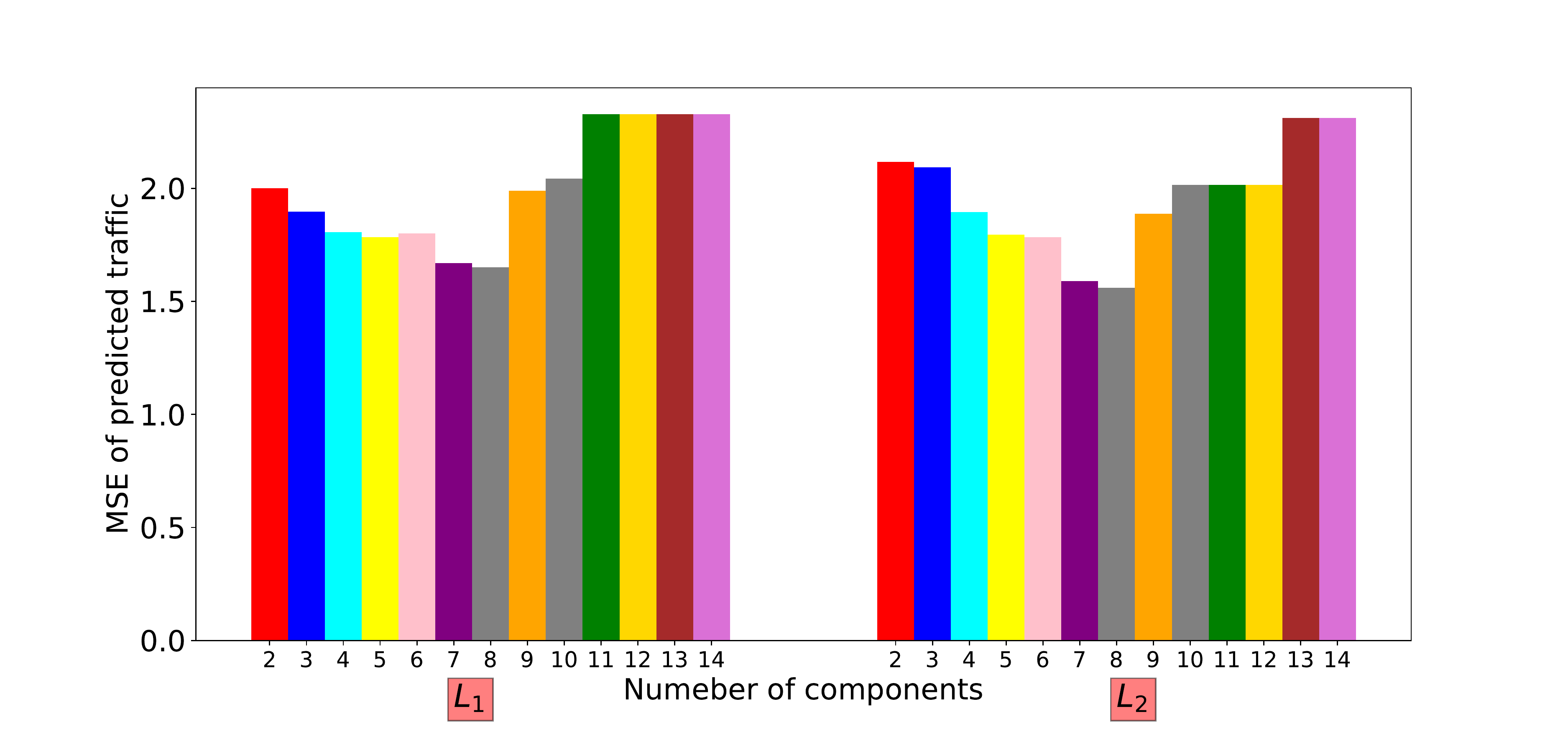}  
  \caption{CDF\_GMM}
  \label{fig:h10}
\end{subfigure}
\vspace{-10pt}
\caption{The impact of the number of similar data centers on the forecasting error in CDF\_eros and CDF\_GMM methods. Models trained on the most similar data centers to the target data center may not always provide the best predictions.}
\label{fig:knum}
\end{figure*}

\begin{table}[ht]
    \caption{Comparison between the forecasting of different methods for $\mathbf{H \in \{10,20\}}$ and $\mathbf{U=12}$.}
    \label{table:forecasting}
    \centering
    \resizebox{\linewidth}{!}{%
        \begin{tabular}{|c|c|c|c|c|c|c|c|c|c|}
            \cline{3-8}
            \multicolumn{2}{c|}{}& \multicolumn{3}{c|}{H=10} & \multicolumn{3}{c|}{H=20}\\
            \hline
            \textbf{Data center}&Metric&\textbf{LSTM}&\textbf{LSTM+GNN}&\textbf{CDF}&\textbf{LSTM}&\textbf{LSTM+GNN}&\textbf{CDF}\\\hline
            \multirow{3}{*}{$L_{1}$}&MSE&0.088&0.039&\textbf{0.012}&0.21&0.03&\textbf{0.001}\\ &MAE&0.2&0.165&\textbf{0.08}&0.4&0.1&\textbf{0.03}\\ &MAPE&0.91&0.7&\textbf{0.52}&0.18&0.1&\textbf{0.07}
            \\\hline

            \multirow{3}{*}{$L_{2}$}&MSE&3.88&3.54&\textbf{3.1}&0.85&0.41&\textbf{0.32}\\ &MAE&1.43&1.35&\textbf{1.27}&0.89&0.63&\textbf{0.57}\\ &MAPE&3.34&2.41&\textbf{2.27}&1.19&0.94&\textbf{0.66}
            \\\hline

            \multirow{3}{*}{$L_{3}$}&MSE&9.6&8.69&\textbf{7.65}&0.35&0.13&\textbf{0.024}\\ &MAE&2.19&2.1&\textbf{1.87}&0.47&0.16&\textbf{0.95}\\ &MAPE&2.08&1.48&\textbf{1.21}&4.23&1.00&\textbf{0.95}
            \\\hline
       
       \multirow{3}{*}{$L_{4}$}&MSE&0.28&0.18&\textbf{0.11}&0.26&0.15&\textbf{0.008}\\ &MAE&0.59&0.47&\textbf{0.27}&0.41&0.18&\textbf{0.079}\\ &MAPE&4.68&2.67&\textbf{2.23}&16.49&8.87&\textbf{3.89}
       \\\hline
       
\multirow{3}{*}{$L_{5}$}&MSE&1.84&1.69&\textbf{1.46}&6.02&3.28&\textbf{2.82}\\ &MAE&0.99&0.96&\textbf{0.88}&1.52&1.26&\textbf{1.00}\\ &MAPE&2.04&1.05&\textbf{1.10}&0.68&0.60&\textbf{0.35}
       \\\hline
       
       \multirow{3}{*}{$L_{6}$}&MSE&1.38&1.21&\textbf{1.12}&4.2&0.028&\textbf{0.004}\\ &MAE&0.79&0.76&\textbf{0.69}&0.46&0.169&\textbf{0.058}\\ &MAPE&2.56&2.16&\textbf{1.65}&2.05&0.72&\textbf{0.25}
       \\\hline
       
       \multirow{3}{*}{$L_{7}$}&MSE&1.48&1.18&\textbf{0.98}&4.21&3.16&\textbf{2.82}\\ &MAE&0.95&0.83&\textbf{0.77}&2.03&1.72&\textbf{1.68}\\ &MAPE&2.21&1.78&\textbf{1.30}&1.09&0.92&\textbf{0.6}
       \\\hline
       
       \multirow{3}{*}{$L_{8}$}&MSE&1.33&1.02&\textbf{0.82}&0.21&0.01&\textbf{0.001}\\ &MAE&0.71&0.64&\textbf{0.57}&0.42&0.08&\textbf{0.017}\\ &MAPE&2.00&1.5&\textbf{1.17}&3.53&0.33&\textbf{0.87}
       \\\hline
       
       \multirow{3}{*}{$L_{9}$}&MSE&1.17&0.73&\textbf{0.47}&0.58&0.21&\textbf{0.062}\\ &MAE&0.95&0.68&\textbf{0.57}&0.75&0.24&\textbf{0.11}\\ &MAPE&3.34&2.42&\textbf{1.60}&1.71&0.54&\textbf{0.38}
       \\\hline
       
       \multirow{3}{*}{$L_{10}$}&MSE&1.88&1.69&\textbf{1.34}&11.91&9.04&\textbf{7.97}\\ &MAE&0.85&0.65&\textbf{0.44}&2.94&2.63&\textbf{2.3}\\ &MAPE&2.17&1.78&\textbf{1.39}&1.29&1.05&\textbf{0.86}
       \\\hline
       
       \multirow{3}{*}{$L_{11}$}&MSE&7.71&6.88&\textbf{6.37}&3.41&1.93&\textbf{1.31}\\ &MAE&2.5&2.36&\textbf{2.26}&1.61&1.26&\textbf{1.07}\\ &MAPE&1.07&0.99&\textbf{0.94}&1.04&0.99&\textbf{0.75}
       \\\hline
       
       \multirow{3}{*}{$L_{12}$}&MSE&0.93&0.74&\textbf{0.6}&2.5&1.005&\textbf{1.31}\\ &MAE&0.73&0.66&\textbf{0.58}&1.53&1.05&\textbf{0.84}\\ &MAPE&2.63&2.28&\textbf{1.39}&1.19&0.98&\textbf{0.78}
       \\\hline
       
       \multirow{3}{*}{$L_{13}$}&MSE&2.05&1.69&\textbf{1.53}&10.05&6.89&\textbf{5.98}\\ &MAE&1.01&0.94&\textbf{0.82}&3.15&2.61&\textbf{2.13}\\ &MAPE&2.41&2.19&\textbf{1.08}&1.09&0.98&\textbf{0.81}
       \\\hline
       
       \multirow{3}{*}{$L_{14}$}&MSE&2.275&1.14&\textbf{0.94}&0.055&0.016&\textbf{0.014}\\ &MAE&1.12&0.88&\textbf{0.79}&0.22&0.1&\textbf{0.08}\\ &MAPE&2.76&1.14&\textbf{1.02}&2.65&0.68&\textbf{0.47}
       \\\hline
       
       \multirow{3}{*}
{$L_{15}$}&MSE&3.83&3.78&\textbf{3.32}&0.75&0.24&\textbf{0.14}\\ &MAE&1.66&1.56&\textbf{1.49}&0.83&0.49&\textbf{0.39}\\ &MAPE&1.92&1.32&\textbf{0.93}&1.27&0.75&\textbf{0.55}
       \\\hline
    \end{tabular}
    }
\end{table}

\subsection{Baselines}
We compare the performance of our framework with different forecasting models:
\begin{itemize}
    \item LSTM: For each data center, we train a forecasting model comprising an LSTM layer and a dense layer using the multivariate time series dataset of the data center.
    \item GNN+LSTM: This approach is similar to the CDF model but we assume that the causal effect of each variable on all other variables is equivalent to 1 ($ \forall m_{i,j} \in M$, $m_{i,j}=1$). 
    \item CDF: This is our proposed forecasting framework represented in Fig. \ref{fig:diag}.
    \item CDF\_GMM: This method is considered as a variant of the Cold-CDF framework.
    In this model, GMM is applied in the cold-start forecasting component of the CDF-cold framework to find the most similar data centers to the target data center with the cold-start problem. 
    \item CDF\_GMM\_sd: This approach is a variant of CDF\_GMM method. In this technique, GMM is used to identify the most similar data centers to the target data center. Then, the CDF models trained on similar data centers are used to forecast the future values of the time series in the target data center. We iteratively remove the forecasts which are out of the range of standard deviation of the remaining forecasts. Finally, we take the average of the remaining forecasts as the predicted values for the target data center.
    \item CDF\_eros: In this approach, the Eros method is used to measure the similarities between different multivariate time series in the cold-start forecasting component of the CDF-cold framework.
    \item CDF\_virtual: 
    In this approach, GMM is used to identify k most similar data centers to the target data center with the cold-start problem. Then, a virtual data center is created by measuring the pointwise average of the time series of similar data centers. Finally, a CDF model is trained using the virtual data center and exploited to forecast in the target data center.

    \item CDF\_virtual\_mn: This model is a variant of CDF\_virtual. In this approach, a virtual data center is created by measuring the pointwise average of the time series of similar data centers weighted by the Manhattan distance of the target data center and each similar data center.
\end{itemize}

\begin{table*}[t]
    \caption{Comparison between the performance of different methods in mitigating cold-start forecasting problem for H=10 and U=12. The best results are highlighted in bold. CDF\_GMM\_sd outperforms other methods in forecasting for all data centers.}
    \label{table:cold_forecasting}
    \centering
    \resizebox{0.9\linewidth}{!}{%
        \begin{tabular}{|c|c|c|c|c|c|c|c|c|c|c|}
            \hline
            \textbf{Data center} & Metric & \textbf{LSTM} & \textbf{LSTM+GNN} & \textbf{CDF} & \textbf{CDF\_eros} & \textbf{CSF\_GMM} & \textbf{CDF\_GMM\_sd} & \textbf{CDF\_virtual} & \textbf{CDF\_virtual\_mn} \\
            \hline
            \multirow{3}{*}{$L_{1}$} & MSE & 2.14 & 1.98 & 1.85 & 1.76 & 1.64 & \textbf{1.41} & 1.69 & 1.66 \\
             & MAE & 1.282 & 1.12 & 0.99 & 0.98 & 0.95 & \textbf{0.9} & 0.98 & 0.98 \\
             & MAPE & 1.63 & 1.35 & 1.24 & 1.07 & 0.98 & \textbf{0.91} & 1.4 & 1.18 \\
             \hline
             
            \multirow{3}{*}{$L_{2}$} & MSE & 2.34 & 1.98 & 1.75 & 1.61 & 1.55 & \textbf{1.40} & 1.64 & 1.62 \\
             & MAE & 0.99 & 0.91 & 0.86 & 0.82 & 0.79 & \textbf{0.77} & 0.85 & 0.82 \\
             & MAPE & 2.99 & 2.21 & 2.07 & 1.86 & 1.71 & \textbf{1.63} & 1.9 & 1.81 \\
             \hline
             
            \multirow{3}{*}{$L_{3}$} & MSE & 3.11 & 2.98 & 2.71 & 2.68 & 2.53 & \textbf{2.31} & 2.63 & 2.56 \\
             & MAE & 1.22 & 1.14 & 1.08 & 1.04 & 1.01 & \textbf{1.01} & 1.04 & 1.02 \\
             & MAPE & 1.87 & 1.51 & 1.49 & 1.47 & 1.19 & \textbf{1.12} & 1.59 & 1.51 \\
             \hline

       \multirow{3}{*}{$L_{4}$}&MSE&0.42&0.28&0.2&0.17&0.12&\textbf{0.09}&0.18&0.16\\ &MAE&0.59&0.41&0.31&0.28&0.25&\textbf{0.24}&0.29&0.27\\ &MAPE&10.67&2.61&2.12&1.88&1.51&\textbf{1.50}&1.83&1.78
       \\\hline
       
\multirow{3}{*}{$L_{5}$}&MSE&1.52&1.32&1.08&0.91&0.86&\textbf{0.73}&0.9&0.87\\ &MAE&0.96&0.81&0.69&0.68&0.66&\textbf{0.56}&0.72&0.69\\ &MAPE&2.31&1.91&1.54&1.32&1.21&\textbf{1.12}&1.78&1.56
       \\\hline
       
       \multirow{3}{*}{$L_{6}$}&MSE&0.89&0.78&0.67&0.61&0.5&\textbf{0.37}&0.6&0.53\\ &MAE&0.76&0.61&0.53&0.49&0.42&\textbf{0.41}&0.5&0.49\\ &MAPE&3.12&1.78&1.52&1.48&1.30&\textbf{1.26}&1.37&1.31
       \\\hline
       
       \multirow{3}{*}{$L_{7}$}&MSE&3.24&2.97&2.51&2.42&2.33&\textbf{2.18}&2.49&2.38\\ &MAE&1.43&1.21&1.09&1.03&1.01&\textbf{0.99}&1.02&1.00\\ &MAPE&1.65&1.35&1.21&1.12&1.02&\textbf{0.95}&1.2&1.14
       \\\hline
       
       \multirow{3}{*}{$L_{8}$}&MSE&1.29&0.98&0.77&0.59&.43&\textbf{0.31}&0.58&0.55\\ &MAE&0.68&0.54&0.49&0.48&0.43&\textbf{0.40}&0.53&0.48\\ &MAPE&2.92&1.96&1.84&1.52&1.45&\textbf{1.38}&2.69&2.13
       \\\hline
       
       \multirow{3}{*}{$L_{9}$}&MSE&0.99&0.88&0.71&0.65&0.51&\textbf{0.40}&0.67&0.64\\ &MAE&0.83&0.71&0.59&0.53&0.52&\textbf{0.51}&0.53&0.52\\ &MAPE&2.77&2.32&1.98&1.63&1.39&\textbf{1.38}&2.10&2.00
       \\\hline
       
       \multirow{3}{*}{$L_{10}$}&MSE&0.86&0.75&0.61&0.53&0.41&\textbf{0.34}&0.58&0.54\\ &MAE&0.69&0.58&0.46&0.4&0.38&\textbf{0.37}&0.42&0.40\\ &MAPE&3.71&2.78&2.43&2.29&2.15&\textbf{1.69}&2.75&2.21
       \\\hline
       
       \multirow{3}{*}{$L_{11}$}&MSE&3.67&2.98&2.74&2.69&2.42&\textbf{2.11}&2.72&2.69\\ &MAE&1.78&1.37&1.33&1.28&1.19&\textbf{1.02}&1.33&1.29\\ &MAPE&1.83&1.54&1.50&1.21&1.07&\textbf{1.00}&1.16&1.08
       \\\hline
       
       \multirow{3}{*}{$L_{12}$}&MSE&1.93&1.63&1.58&1.53&1.42&\textbf{1.19}&1.59&1.51\\ &MAE&1.02&0.96&0.86&0.8&0.76&\textbf{0.71}&0.9&0.78\\ &MAPE&3.72&2.89&2.54&1.47&1.38&\textbf{1.21}&1.65&1.40
       \\\hline
       
       \multirow{3}{*}{$L_{13}$}&MSE&2.53&2.02&1.81&1.64&1.51&\textbf{1.28}&1.98&1.78\\ &MAE&1.16&1.03&0.99&0.98&0.95&\textbf{0.89}&1.04&1.01\\ &MAPE&6.52&5.67&4.22&3.90&3.62&\textbf{2.65}&10.64&7.04
       \\\hline
       
       \multirow{3}{*}{$L_{14}$}&MSE&0.98&0.89&0.76&0.71&0.61&\textbf{0.50}&0.87&0.7\\ &MAE&0.83&0.75&0.96&0.84&0.68&\textbf{0.68}&0.75&0.69\\ &MAPE&1.84&1.70&1.56&1.22&1.07&\textbf{1.02}&1.8&1.65
       \\\hline
       
       \multirow{3}{*}{$L_{15}$}&MSE&4.21&3.93&3.76&3.21&3.05&\textbf{2.82}&3.76&3.43\\ &MAE&1.52&1.43&1.31&1.3&1.24&\textbf{1.21}&1.37&1.27\\ &MAPE&1.62&1.58&1.42&1.21&1.09&\textbf{1.01}&2.44&1.16
       \\\hline
       
    \end{tabular}
    }
\end{table*}
\subsection{Experimental Setup}
To smooth time series data over outliers and short-term fluctuations, we use the rolling median technique in which
the attribute values of a sliding time window are replaced with the median of the values in that window. In this paper, we set the window size to 7.
We utilize first-order differencing ($x^j_{i,t} \leftarrow x^j_{i,t}-x^j_{i,t-1}$) to convert non-stationary datasets to stationary. In non-stationary time series datasets, the statistical properties of the dataset ( e.g., mean and variance) change in time. 
To standardize the dataset, we use Z-score normalization (Eq. \ref{zstarnd}) in which attributes are rescaled to ensure the mean and the standard deviation are 0 and 1, respectively. Each time series is standardized as:
\begin{align}
    \hat{x}^j_{i,T} = \frac{x^j_{i,T}-\mu_j}{\sigma_j},
    \label{zstarnd}
\end{align}
where $\mu_j$ and $\sigma_j$ are the mean and standard deviation of time series $a_j$. We conduct data standardization only in CDF component of our model. In the cold-start forecasting component, we use non-standardized datasets to preserve the distribution.

We vary hyper-parameters for each baseline method and each dataset to achieve their best performance on this task. We split each dataset into $80\%$ training, $10\%$ validation, and $10\%$ test datasets. 
To train LSTM, GNN and CDF models, we search the learning rate in $\{10^{-1}, 10^{-2}$ $, 10^{-3}, 10^{-4} \}$, the number of epochs in $\{10, 20, 30, 50, 70, 100 \}$ and the batch size in $\{16,32,64,128\}$. The number of hidden units is chosen from $\{10,20,100,200,300\}$ and the number of hidden layers is set from $\{1,2,3\}$.
We set the observable past window size $U=10$ and horizon $H \in \{1,10\}$. We use Mean Squares Error (MSE) loss function and
the RMSProp optimizer to optimize the parameters of our model. 
For causal discovery, we exploit VARLiNGAM \cite{hyvarinen-mlr10} which is an extension of the LiNGAM \cite{shimizu-mlr06} model to time series datasets. VARLiNGAM enables analyzing both lagged and contemporaneous (instantaneous) causal relations in multivariate time series datasets. 

To evaluate the effectiveness of various forecasting models in datasets with the cold-start problem, we remove the historical data for ten of the Google services with high network traffic until t=400 in each data center. We then predict the total network traffic beyond this time step (H $\in \{10,20\}$) when the removed service is added to the data center. Each time, we select one data center as the dataset with the cold-start problem, we consider the other 14 data centers as potentially similar data centers.

\textbf{Evaluation Metrics}:
To assess the performance of different models, we follow existing literature \cite{li-iclr17,wu-ijcai19} and report the mean absolute error (MAE), mean square error (MSE) and mean absolute percentage error (MAPE) of the total network traffic in each data center.


\commentout{
\begin{figure*}[h]
\begin{subfigure}{1\textwidth}
  \centering
  \includegraphics[width=\textwidth]{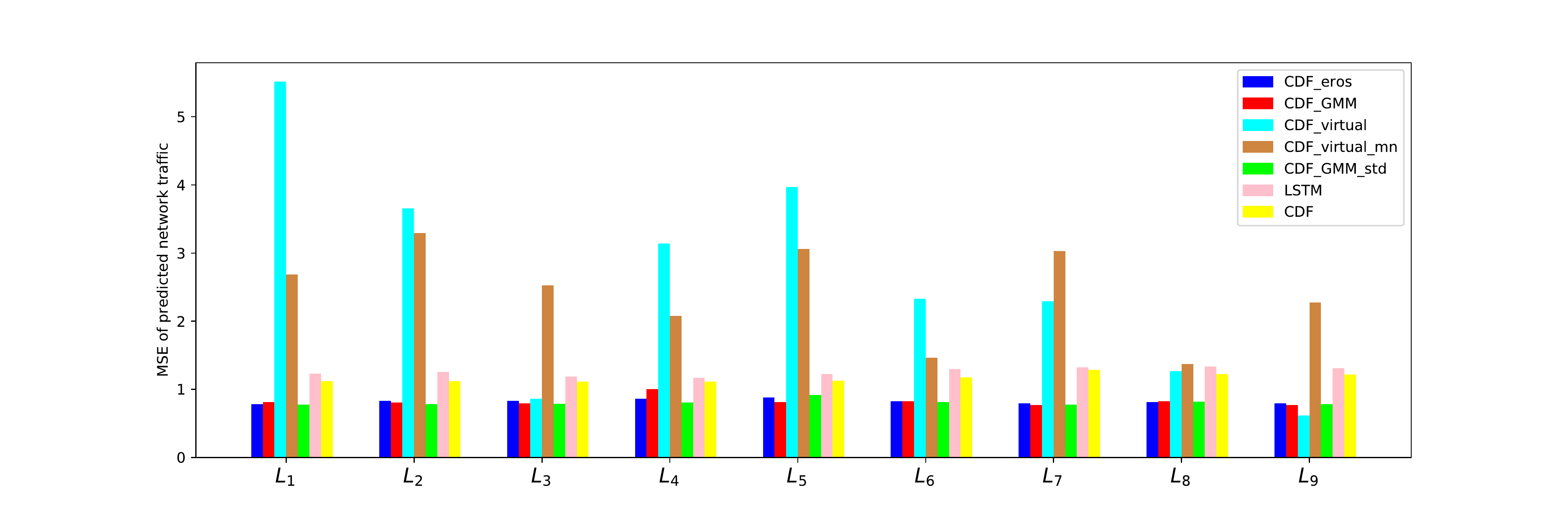}  
  \caption{H=1}
  \label{fig:h1}
\end{subfigure}
\begin{subfigure}{1\textwidth}
  \centering
  \includegraphics[width=\textwidth]{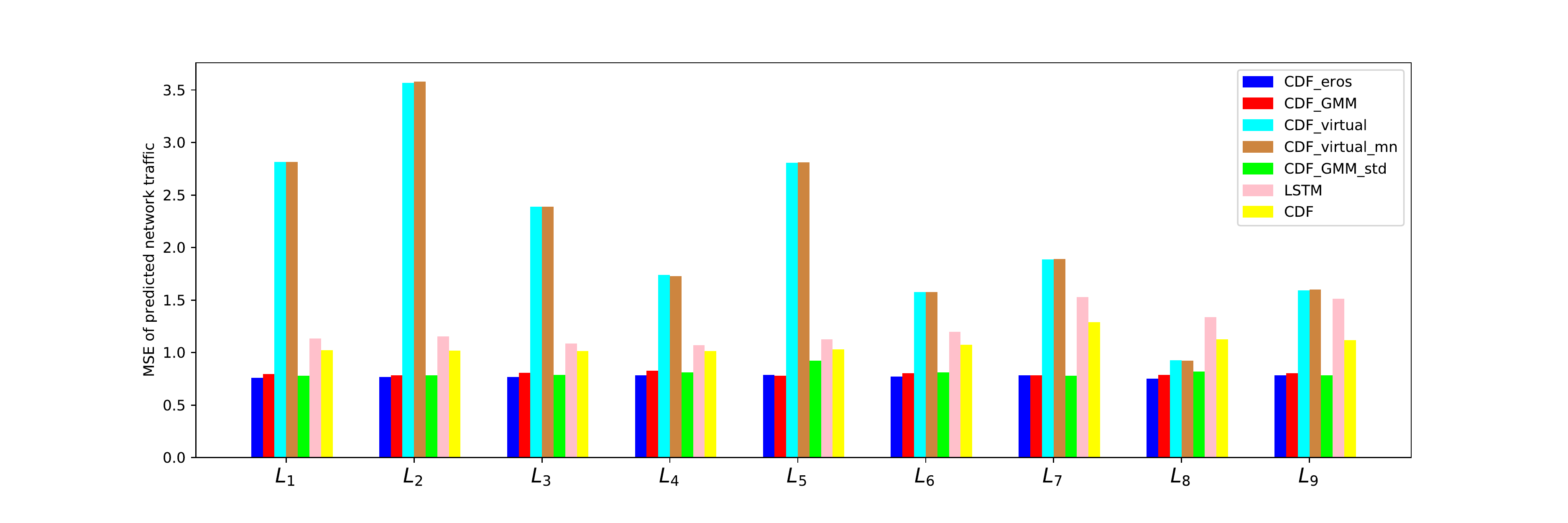}  
  \caption{H=10}
  \label{fig:h10}
\end{subfigure}
\caption{Comparison between the performance of different methods in time series forecasting with the cold-start problem.}
\end{figure*}
}
\subsection{Results}
\textbf{Sensitivity to the number of similar data centers}:
In this experiment, our objective is to explore how the number of similar data centers affects the performance of different cold-start forecasting component variants. As shown in Fig. \ref{fig:knum}, our findings demonstrate that the model trained exclusively on the most similar data center does not necessarily yield the most precise forecasting for data centers that encounter the cold-start problem. In the CDF\_eros method, averaging the forecast of the models trained on the first five most similar data centers produces the least forecasting error.

Regarding the CDF\_GMM method, we observe that partitioning the dataset into 2 or 3 dense clusters or sparse clusters with 12, 13, and 14 partitions leads to a significantly higher forecasting error. Conversely, configuring the number of components to 7 or 8 results in the lowest prediction error.

\textbf{Evaluating the forecasting performance}:
In this experiment, the performance of three different models in total network traffic forecasting was compared. Table \ref{table:forecasting} presents the MSE, MAE, and MAPE of total network traffic predicted by three different methods in 15 different Google data centers. The results show that applying GNN can significantly improve the accuracy of forecasting compared to LTSM in all data centers. For example, LSTM+GNN reduces the MSE of forecasting by $55.5\%$ in $L_{1}$, $9.14\%$ in $L_3$, $35.7\%$ in $L_4$, $23.3\%$ in $L_8$, and $37.6\%$ in $L_9$ for H=10.
Compared to LSTM+GNN, CDF decreases the error from $0.039$ to $0.012$ in $L_{1}$, from $8.69$ to $7.65$ in $L_3$, from $1.18$ to $0.98$ in $L_4$, and from $1.14$ to $0.94$ in $L_{5}$ for $H=10$.

To study the ability to forecast longer horizons, we increase the forecasting horizon from $H=10$ to $H=20$.
Similar to the results when $H=10$, GMM outperforms the LSTM model in reducing MSE ($85.7\%$ in $L_{1}$, $62.8\%$ in $L_{3}$, $42.3\%$ in $L_4$, $63.7\%$ in $L_9$).
Furthermore, we observe that CDF with a causal component consistently outperforms LSTM+GNN in all datasets (e.g., $0.03$ vs. $0.001$ in $L_{1}$, $0.13$ to $0.024$ in $L_3$, from $0.15$ to $0.004$ in $L_4$, and from $3.28$ to $2.82$ in $L_{5}$) for H=20.
In $L_{8}$ and $L_{15}$ the differences are more subtle than in the other data centers with $95\%$ and $68\%$ error reduction, respectively. 
This finding aligns with our hypothesis that the incorporation of causal relationships between various variables and the target variable significantly enhances the accuracy of forecasting. 

\textbf{Cold-start forecasting evaluation}:
To evaluate different methods for mitigating the cold-start problem in multivariate time series forecasting, we measure the mean squared error (MSE), mean absolute error (MAE), and mean absolute percentage error (MAPE) of total network traffic prediction using the forecasting methods. As depicted in Table \ref{table:cold_forecasting}, the methods that lack a cold-start forecasting component (LSTM, LSTM+GNN, and CDF) exhibit higher estimation error compared to the methods that include a cold-start forecasting component. This can be attributed to the fact that LSTM cannot, LSTM+GMM, and CDF cannot learn the interdependence between variables and the total network traffic without historical data for some variables. However, we observed that CDF outperforms LSTM and LSTM+GNN in all datasets which is consistent with the results reported in Table. \ref{table:forecasting}.

Among the Cold-CDF variants, cluster-based methods (i.e., CDF\-\_GMM, and CDF\_GMM\_sd) exhibit the least forecasting error. Compared to CDF, the CDF\_GMM\_sd approach improves the MSE by $23.7\%$ in $L_1\%$, $67.8\%$ in $L_4$, $59.7\%$ in $L_8$, and $29.2\%$ in $L_13$. The findings for the other two metrics (MAE and APE) are consistent with these results. CDF\_GMM and CDF\_GMM\_sd also demonstrate the lowest values for MAE and MAPE.

%% file: 6-conclusion.tex
\section{Conclusion}
\label{conclusion}
In this paper, we present the Cold Causal Demand Forecasting framework, a novel approach designed to address the challenge of cold-start forecasting in multivariate time series data where some variables lack historical information. Our framework leverages causal discovery algorithms to uncover cause-and-effect relationships among interdependent variables. These discovered relationships are then integrated into a neural network model, resulting in improved forecasting accuracy.
To tackle the cold-start problem, we propose a similarity-based approach for forecasting multivariate time series data that lack prior historical data for some variables. 
 Our comprehensive evaluation of 15 Google multivariate time series datasets reveals the superior performance of our framework in comparison to existing baseline methods.
One potential future direction is to integrate transformers for causal inference and propose a model that can forecast in datasets with no historical data for all variables.